\documentclass{ecai}
\usepackage{times}
\usepackage{graphicx}
\usepackage{latexsym}
\usepackage{url}
\usepackage{hyperref}
\usepackage{amsmath}
\usepackage{amssymb}

\ecaisubmission

\begin{document}

\title{Warped Input Gaussian Processes\\for Time Series Forecasting}

\author{David Tolpin \institute{Ben-Gurion University of the
Negev \& PUB+, Israel, email: david.tolpin@gmail.com}}

\maketitle
\bibliographystyle{ecai}

\begin{abstract}
We introduce a Gaussian process-based model for handling of
non-stationarity. The warping is achieved
non-parametrically, through imposing a prior on the relative
change of distance between subsequent observation inputs.
The model allows the use of general gradient optimization
algorithms for training and incurs only a small
computational overhead on training and prediction. The model
finds its applications in forecasting in non-stationary time
series with either gradually varying volatility, presence of
change points, or a combination thereof.  We evaluate the
model on synthetic and real-world time series data comparing
against both baseline and known state-of-the-art approaches
and show that the model exhibits state-of-the-art
forecasting performance at a lower implementation and
computation cost.
\end{abstract}

\section{INTRODUCTION}

Gaussian processes~\cite{RW05} possess properties that make
them the approach of choice in time series forecasting:
\begin{itemize}
    \item A Gaussian process works with as little or as much
    	data as available.
    \item Non-uniformly sampled observations, missing
    	observations, and observation noise are handled
    	organically.
    \item Uncertainty of a future observation is predicted along
		with the mean.
\end{itemize}
In a basic setting though, a Gaussian process  models a
stationary time series with homoscedastic noise.  When either
the covariance between observations or the noise vary depending
on observations inputs or outputs, predictions produced by a
Gaussian process with a stationary kernel and constant noise
variance will be either biased or overly uncertain, hence
handling non-stationarity and heteroscedasticity is crucial in
many applications. Non-stationarity often arises in financial
time series, where market volatility, affecting forecasting
uncertainty, changes with time~\cite{WHG14,NB18};
heteroscedastic noise is common in vital signs monitoring of
patients in intensive care units, where the noise depends on
patient activity and medical interventions~\cite{CRC19}, both
non-stationarity and heteroscedasticity are characteristic for
time series of sensor readings in mobile robotics~\cite{KPB07}.

Various approaches have been proposed for handling
non-stationarity using Gaussian processes. When a time series is
piecewise stationary, change point detection is deemed an
appropriate model, with  a stationary homoscedastic Gaussian
process modelling stretches between consequent change
points~\cite{GOR09,STR10,CV11}.  In cases where the covariance
or noise change gradually and smoothly, it is common to
introduce a non-parametric dependency of kernel and noise
parameters on inputs~\cite{GWB98,PS03,LSC05,SG06,KPB07,WN12},
however, this makes structural modelling of time series, which
constitutes an important advantage of Gaussian processes and
facilitates introduction of prior knowledge in the model, more
challenging.

Another popular way to handle both abrupt and gradual changes in
time series is through mapping of the input
space~\cite{SG92,MK98,DL13,SSZ+15,BHH+16,MG18,CPR+16}.  Covariance
between observations depends on observations inputs as well as
on kernel parameters, and non-stationarity can be modelled by
smoothly modifying observation inputs (warping the input space).
Several methods have been proposed to learn the input space
transformation, and a number of other approaches can be viewed
as employing input space warping for handling non-stationarity.
However, many such approaches either meet difficulties in
practical application, or require elaborated inference
algorithms~\cite{TBT15,BHH+16,SD17,D18}, which may impact the
simplicity of use of Gaussian processes.

In this work we introduce a model for non-parametric warping of
input space for Gaussian processes, suitable in particular for
time series forecasting but also applicable to other domains.
The model is easy to implement, imposes only a small
computational overhead on training and prediction, and allows to
use the whole arsenal of Gaussian process kernels to model time
series structure using prior knowledge. We provide a reference
implementation of the model and evaluate the model on a
synthetic and real-world data, comparing forecasting performance
with both baseline and state-of-the-art approaches and show that
the model exhibits state-of-the-art forecasting performance at a
lower implementation and computation cost.

This work brings the following contributions:
\begin{itemize}
	\item A novel approach to handling non-stationarity
		in Gaussian processes.
	\item A Gaussian process model for forecasting in
		non-stationary time series.
	\item A reference implementation of the model within a
		probabilistic programming framework.
\end{itemize}

\section{PRELIMINARIES}

A Gaussian Process is a collection of random variables, any
finite number of which have (consistent) joint Gaussian
distributions.  A Gaussian process is fully specified by its
mean function $m(x)$ and covariance, or kernel, function
$k(x,x')$ and defines a distribution over functions. The mean
function is often set to zero, $m(x) \equiv 0$.  A Gaussian
process defines a distribution over functions:
\begin{equation}
	 f \sim \mathcal{GP}(m(x), k(x,x'))
\end{equation}
Any finite set of values of function $f$ at inputs $\pmb{x}$
follows the multivariate normal distribution
$\mathcal{N}(\pmb{\mu_x}, \Sigma_{\pmb{x}})$ with mean
$\pmb{\mu_x} = m(\pmb{x})$ and covariance matrix $\Sigma_{\pmb{x}} = \{k(x_i,x_j)\}$.

Posterior inference in Gaussian processes can be performed
analytically. Let $\pmb{f}$ be the observations at inputs 
$\pmb{x}$. Then the posterior distribution of values $\pmb{f}_*$ 
at inputs $\pmb{x}_*$ is
\begin{equation}
	\pmb{f}_*|\pmb{f} \sim \mathcal{N}(\pmb{\mu_{x_*}} + \Sigma_{\pmb{xx_*}}^\top\Sigma^{-1}(\pmb{f}-\pmb{\mu_{x}}), \Sigma_{\pmb{x_*}} - \Sigma_{\pmb{xx_*}}^\top\Sigma^{-1}\Sigma_{\pmb{xx_*}})
	\label{eq:posterior}
\end{equation}
where $\Sigma_{\pmb{xx_*}}$ is the covariance matrix between $\pmb{x}$ and
$\pmb{x_*}$.

Kernel functions normally have hyperparameters; we shall write
$k(x, x'; \theta)$ to denote that the kernel function $x$ has
hyperparameter $\theta$, possibly multidimensional, or omit the
hyperparameters when they are clear from the context. Training a
Gaussian process involves choosing $\theta$ based on the
observations. For example, the Gaussian, or RBF, kernel has
the form
\begin{equation}
	\mathrm{RBF}(x, x'; l) = \exp\left(- \frac {(x-x')^2} {2l^2}\right)
	\label{eq:rbf}
\end{equation}
and is parameterized by a single hyperparameter $l$.

A straightforward way to choose $\theta$ is to maximize
log marginal likelihood $L$ of observations $(\pmb{x}, \pmb{f})$:
\begin{equation}
	L = \log p(\pmb{f}|\pmb{x},\theta) = - \frac 1 2 | \Sigma |
	- \frac 1 2(\pmb{f} - \pmb{\mu})^\top\Sigma^{-1}(\pmb{f} -
	\pmb{\mu}) - \frac n 2 \log (2 \pi)
	\label{eq:lml}
\end{equation}
where $n$ is the number of observations.

There is no closed form solution for maximizing $L$ in general,
however gradient-based optimization methods allow to obtain an
approximate solution efficiently.

\section{WARPED INPUT GAUSSIAN PROCESS MODEL}

A major advantage of Gaussian process regression in general, and
for application to time series in particular, is the explicit
inclusion of uncertainty in the model: both the mean and the
variance are predicted at unobserved inputs. However, perhaps
somewhat counterintuitively, the variance, given the kernel and
the kernel's hyperparameters, does not depend on observed
outputs. Indeed, the covariance matrix in equation
(\ref{eq:posterior}) does not depend on $\pmb{f}$. 

One way to circumvent this limitation of Gaussian processes is
to introduce non-stationarity into the kernel function, such
that the covariance depends on  both the distance between
inputs, $||x, x'||$ and on inputs themselves. In some
kernels, such as the dot product kernel $k(x,x') = x \cdot x'$,
non-stationarity is fixed in the kernel design. In other
kernels, non-stationarity comes through dependency of kernel
hyperparameters on the inputs, and the dependency $\theta(x,x')$
itself can be learned from data~\cite{G97,PS03,MG18}. Related to
varying kernel hyperparameters with inputs is the idea
of \textit{warping} the input space~\cite{SG92}. A stationary
kernel depends on both the distance between inputs and
the hyperparameters. Consider, for example, the RBF kernel
(\ref{eq:rbf}). Increasing hyperparameter $l$, customarily called
the length scale, has the same effect on the covariance as
decreasing the distance between $x$ and $x'$ by the same
relative amount. Moving points away from each other will
effectively decrease the length scale and covariance between the
points. Warping of the input space has an intuitive
interpretation for time series: the time runs faster in areas
with high output volatility and slower when the output is
stable.

A research problem addressed  by different warping methods is
how the warping is represented and what objective should be
maximized to learn the optimal warping for a given problem
instance.  In what follows, we introduce warping of the input space
of a one-dimensional Gaussian process by imposing a prior on
the distances between adjacent inputs. We train the
process by maximizing the combined log marginal likelihood of the
observations under the prior and of the Gaussian process. The
model is trivially extended to a multi-dimensional Gaussian
process where only a single dimension is warped, such a
in the case of a time series where there are multiple predictors
but only the time needs to be warped to account for temporal
non-stationarity.

\subsection{Model}

In a Gaussian process model for handling non-stationarity
through displacement of observation inputs, the choice is of the
form of the prior imposed on the inputs. One option is to impose
a Gaussian process prior on the inputs. This is a rich prior
allowing to model complex structured non-stationarity; deep
Gaussian processes~\cite{D18} is a realization of such prior.
However, inference in the presence of such prior requires
special techniques and is computationally expensive. On the
other extreme is imposing an independent Gaussian prior on each
input, which is related to the modelling of input
uncertainty~\cite{MR11,DTL16}. \cite{MR11} show though that
independent input noise may be reduced to independent output
noise, and as such is not sufficiently expressive for modelling
non-stationarity for forecasting. Here, we propose a prior that
is just a single step away from an independent prior on each
input, namely one which corresponds to a 3-diagonal striped
covariance matrix $\Sigma=\{\sigma_{ij}\}$, such that
$\sigma_{ij}=0 \; \forall |i-j| > 1$, which is equivalent to
imposing independent priors on \textit{distances} between
adjacent inputs. An intuition behind this prior is that the
distance between adjacent locations increases, and the effective
length scale decreases, in areas with high volatility, and vice
versa in areas with low volatility. For convenience of
inference, we formulate the prior in terms of relative change of
distance between inputs. We exploit the structure of this prior
to specify the model compactly, without having to manipulate the
full covariance matrix of the prior.

Formally, let $\mathcal{GP}$ be a one-dimensional Gaussian
process. Let also $D$ be a distribution  on $\mathcal{R}^+$.
Then, given inputs $\pmb{x}, x_{i+1} > x_{i} \forall i$, the
generative model for outputs is
\begin{align}
	\label{eq:gen}
	\tilde{x}_1 & = x_1 \\ \nonumber
	\lambda_i & \sim D \\ \nonumber
	\tilde{x}_i & = \tilde{x}_{i - 1} + \lambda_i(x_i - x_{i - 1}) \mbox{ for } i > 1\\ \nonumber
	\pmb{f} & \sim \mathcal{GP}(\pmb{\mu}_{\pmb{\tilde{x}}}, \Sigma_{\pmb{\tilde{x}}})
\end{align}
In words, inputs $\pmb{x}$ are transformed (warped) into
$\pmb{\tilde{x}}$  by stretching or compressing distances
between adjacent inputs $x_{i-1}$ and $x_{i}$ by relative amounts
$\lambda_i$ drawn from $D$. For brevity, we call the introduced
model WGP in the rest of the paper. $D$ serves as a prior belief
on distances between adjacent inputs, relative to the original
distances. Without loss of generality, the mean of $D$ can be
assumed to be 1, so that the mean of the prior belief is that no
warping is applied. 

\subsection{Training}

Training of a WGP model is performed by maximizing the log
marginal likelihood $L_{WGP}$:

\begin{equation}
	L_{WGP} = L + \sum_{i=2}^{n} \log p_D\left(\frac {\tilde{x}_i - \tilde{x}_{i-1}} {x_i - x_{i-1}}\right) + C
	\label{eq:lw}
\end{equation}
where $C$ is a normalization constant that does not depend on
either hyperparameters or observations and is not required for
training. As with kernel hyperparameters, derivatives of $L_W$
by both hyperparameters and transformed inputs $\pmb{\tilde{x}}$ are
readily obtainable analytically or through algorithmic
differentiation~\cite{GW08}.

\subsection{Forecasting}

After training, forecasting is virtually identical to that of a
regular Gaussian process, with one exception: for prediction in
a new location $x_*$, the warped image $\tilde{x}_*$ of the
location must be obtained for substituting into
(\ref{eq:posterior}). The possible options are:
\begin{itemize}
	\item Choosing $\tilde{x}_*$ that maximizes $L_{WGP}$ for $\pmb{x}\circ x_*$ and $\pmb{f} \circ f*$.
	\item Setting $\lambda_*=1$ and, consequently, $\tilde{x}_* = \tilde{x}_n + x_* - x_n$.
	\item Setting $\lambda_*=\lambda_n$ and $\tilde{x}_* = \tilde{x}_n + (x_* - x_n)\frac {\tilde{x}_n - \tilde{x}_{n-1}} {x_n - x_{n-1}}$.
\end{itemize}

The first option is best aligned with log marginal likelihood
maximization during training but computationally expensive. The
last option expresses a smoothness assumption: the length scale
is likely to be similar in adjacent inputs. We experimented
with the three options and found that empirically on synthetic
and real-world datasets predictive accuracy of the third option
is virtually indistinguishable from the first one. In the
empirical evaluation, we computed the warped location for
forecasting as $\tilde{x}_n + \lambda_n(x_* - x_n)$.

\subsection{Modelling Seasonality}

Time series are often modelled by combining \textit{trend} and
\textit{seasonality}, that is, similarity between nearby
observations on one hand and observations at similar phases of a
period on the other hand. In Gaussian processes, kernels based
on the periodic kernel~\cite{MK98} are used to model seasonality.
Warping of the input space, used to maintain  would interfere
with dependencies induced by the periodic kernel. Consider
monitoring of vital sign time series in intensive care
unit~\cite{CCP+12}: while volatility of the time series may
evolve over time, and warping the time may be adequate for
modelling non-stationarity, observations at the same
\textit{astronomical} time of the day tend to be similar.

A solution for warping the trend time but keeping the
seasonality time unwarped is to include both original and warped
dimension into the input space. This way, kernel features
modelling the trend and thus affected by non-stationarity are
made dependant on the warped time, and those modelling
seasonality --- on the original time. Generative
model~(\ref{eq:gen-seasonality}) extends (\ref{eq:gen}) by
combining $\pmb{x}$ and $\pmb{\tilde{x}}$ on input to the
Gaussian process:
\begin{align}
	\label{eq:gen-seasonality}
	\tilde{x}_1 & = x_1 \\ \nonumber
	\lambda_i & \sim D \\ \nonumber
	\tilde{x}_i & = \tilde{x}_{i - 1} + \lambda_i(x_i - x_{i - 1}) \mbox{ for } i > 1\\ \nonumber
	\pmb{f} & \sim \mathcal{GP}(\pmb{\mu}_{\pmb{\tilde{x} \circ x}}, \Sigma_{\pmb{\tilde{x} \circ x}})
\end{align}

Consider, for example, the following kernel, composed of locally
periodic and trend terms:
\begin{align}
	\label{eq:periodic-trend}
	& k(x, x'; \theta) = c_1 \mathrm{RBF}(x, x') \mathrm{Periodic}(x, x') + c_2 \mathrm{Matern_{\frac 3 2}}(x, x') \\ \nonumber
	& \mbox{where} \\ \nonumber
	& \mathrm{RBF}(x, x'; l_1) = \exp \left( - \frac {(x-x')^2} {2l_1^2}\right) \\ \nonumber
	& \mathrm{Periodic}(x, x';p, l_2) = \exp \left( - \frac {2\sin^2\left(\frac {\pi|x-x'|} p \right)} {l_2^2}\right) \\ \nonumber
	& \mathrm{Matern_{\frac 3 2}}(x, x';l_3) = \left(1 + \frac {\sqrt{3}|x-x'|} {l_3}\right)\exp\left(-\frac {\sqrt{3}|x-x'|} {l_3}\right) \\ \nonumber
	& \theta = (c1, c_2, l_1, l_2, l_3, p)
\end{align}
In this kernel, the $\mathrm{RBF}$ and $\mathrm{Matern_{\frac 3
2}}$ components reflect local dependencies between inputs
and hence should be affected by input warping. The
$\mathrm{Periodic}$ component, however, expresses dependencies
between points at similar phases of different periods, with the
period length $p$ normally known upfront and staying fixed.
Thus, in the presence of warping, the modified kernel
$\tilde{k}(\cdot, \cdot)$ must receive both warped and original
inputs and pass appropriate inputs to each of the
components:
\begin{align}
	\label{eq:periodic-trend-warped}
	&\tilde{k} ((\tilde{x},x), (\tilde{x'},x'); \theta) =  \\ \nonumber
	& \hspace{6em}c_1 \mathrm{RBF}(\tilde{x}, \tilde{x'}) \mathrm{Periodic}(x, x')\!+\!c_2 \mathrm{Matern_{\frac 3 2}}(\tilde{x}, \tilde{x'})
\end{align}

\section{EMPIRICAL EVALUATION}

\begin{table*}
	\centering
	\caption{Negative log predictive density on synthetic datasets.}
	\label{tab:nlpd-synthetic}
	\begin{tabular}{r  c  c  c c}
		{\bf dataset} & {\bf no warping} & {\bf warped} & {\bf warped, periodic} & {\bf deep GP} \\ \hline
		trend & 0.2734$\pm$0.0464 & {\bf 0.2384$\pm$0.0649}& 0.2387$\pm$0.0620 & 0.6110$\pm$0.0511 \\
		trend+seasonal & -0.2575$\pm$0.0273 & -0.3278$\pm$0.0288 & {\bf -0.3683$\pm$0.0312} & 0.1236$\pm$0.0659 \\
	\end{tabular}
\end{table*}

\begin{figure*}
  \begin{minipage}[c]{0.33\textwidth}
    \centering
	  \includegraphics[width=0.96\linewidth]{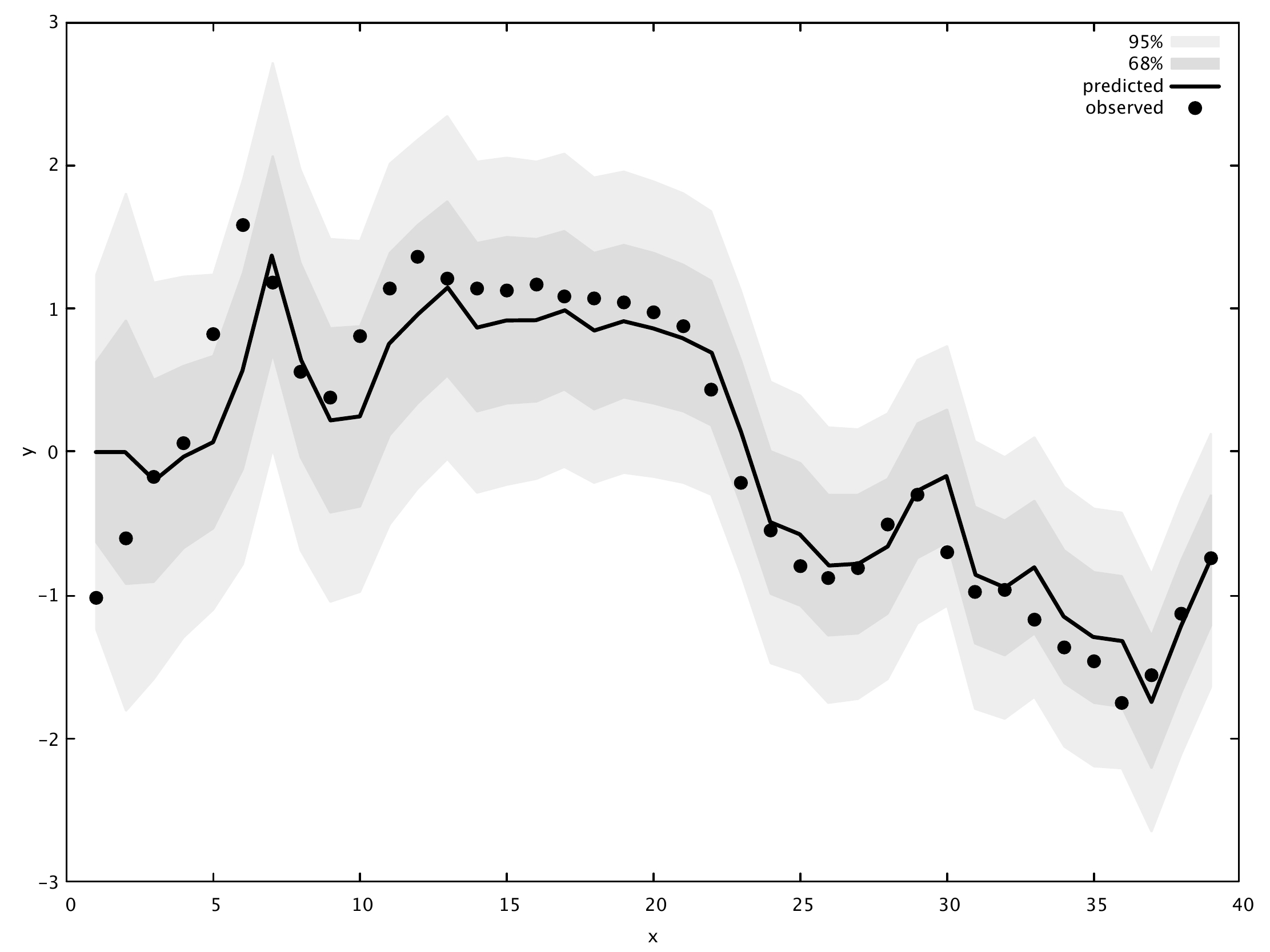}

		(a) no-warping, NLPD = 0.1887
    \end{minipage}
    \begin{minipage}[c]{0.33\textwidth}
    \centering
	\includegraphics[scale=0.5,width=0.96\linewidth]{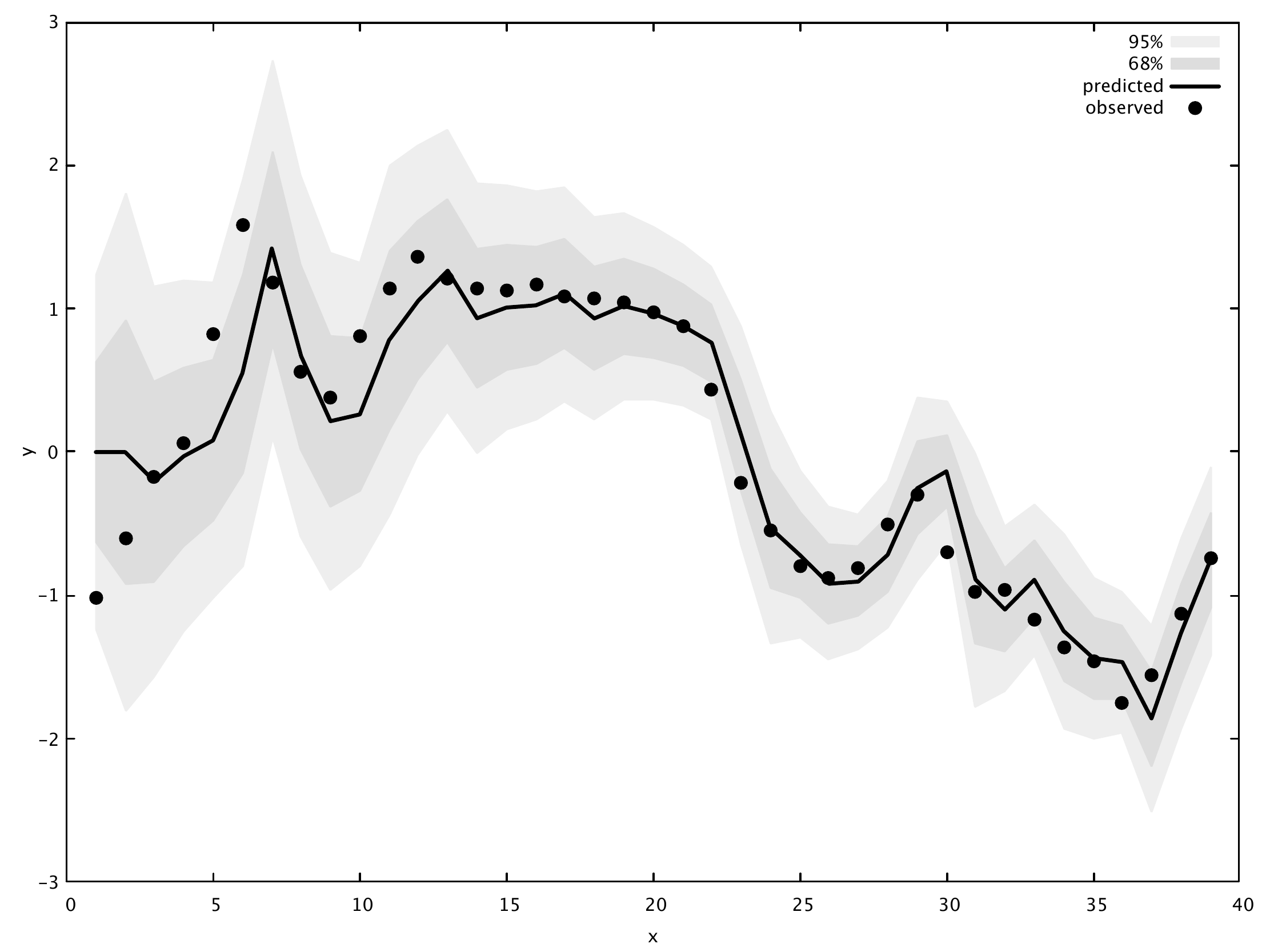}

		(b) warped, NLPD = -0.1620
    \end{minipage}
    \begin{minipage}[c]{0.33\textwidth}
    \centering
	\includegraphics[scale=0.5,width=0.96\linewidth]{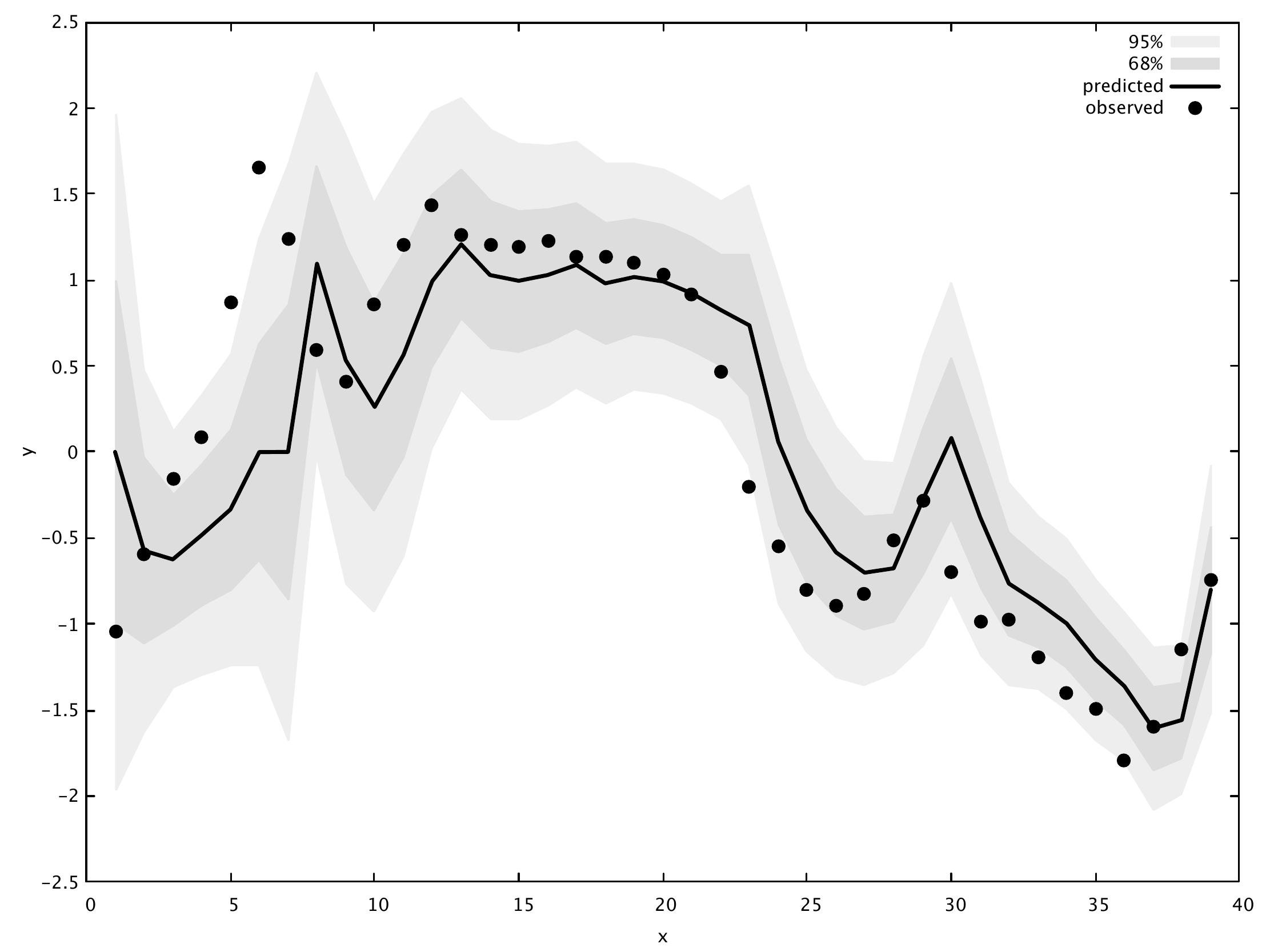}

		(c) deep GP, NLPD = -0.0912
    \end{minipage}
	\caption{Forecasting on an instance from the synthetic dataset.}
	\label{fig:synthetic}
\end{figure*}

\begin{table*}
	\centering
	\caption{Negative log predictive density on real-world datasets.}
	\label{tab:nlpd-real}
	\begin{tabular}{r  c  c  c}
		{\bf dataset} & {\bf no warping} & {\bf warped} & {\bf deep GP} \\ \hline
		LIDAR & 0.2543 & {\bf 0.2290} &  0.2370 \\ 
		Marathon & 0.1887 & -0.1620 & {\bf -0.2183} \\
		Motorcycle & 1.9320 & {\bf 0.8063} & 1.191897 
	\end{tabular}
\end{table*}

\begin{figure*}
  \begin{minipage}[c]{0.33\textwidth}
    \centering
	  \includegraphics[width=0.96\linewidth]{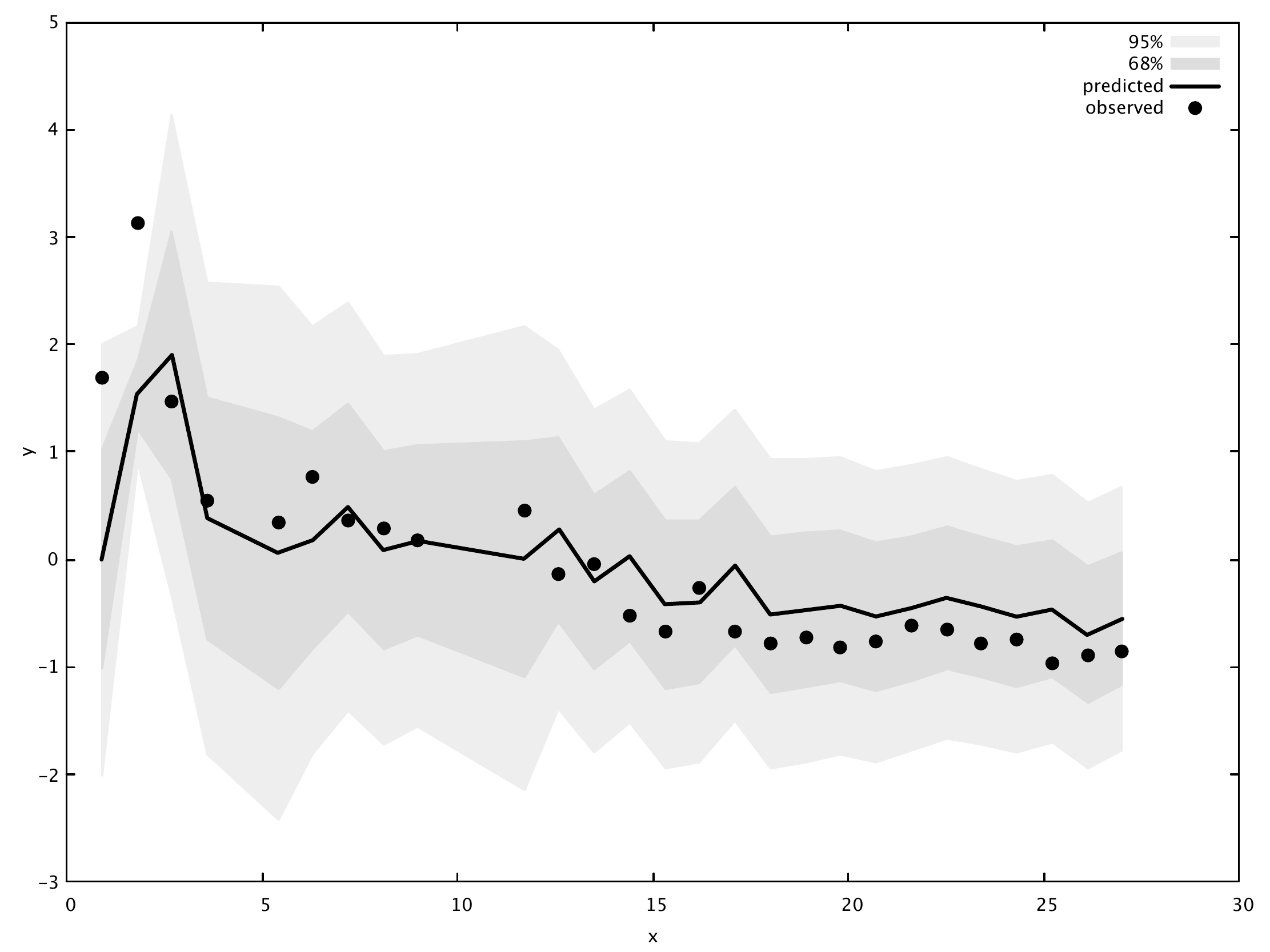}

		(a) no-warping, NLPD = 0.1887
    \end{minipage}
    \begin{minipage}[c]{0.33\textwidth}
    \centering
	\includegraphics[scale=0.5,width=0.96\linewidth]{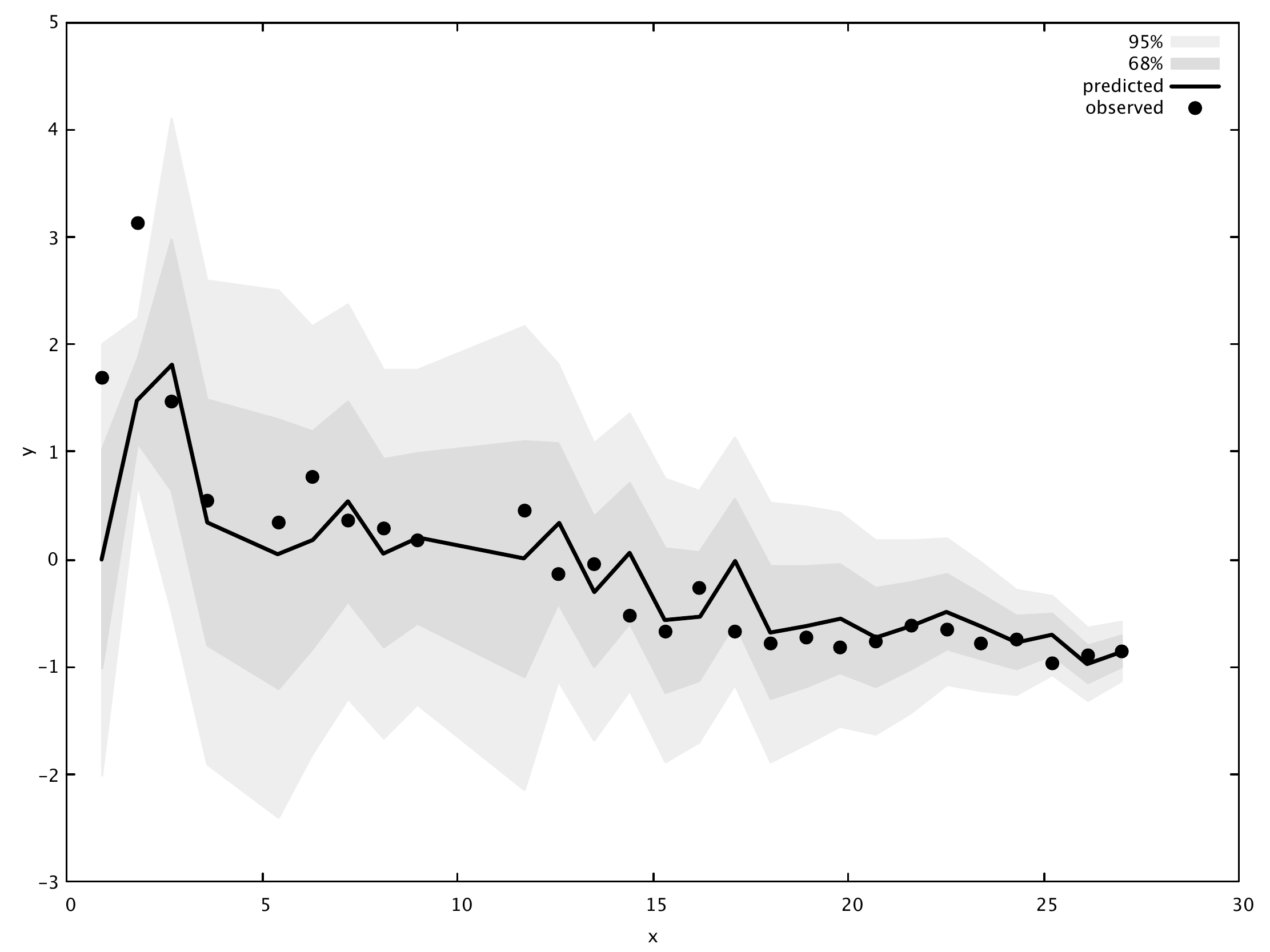}

		(b) warped, NLPD = -0.1620
    \end{minipage}
    \begin{minipage}[c]{0.33\textwidth}
    \centering
	\includegraphics[scale=0.5,width=0.96\linewidth]{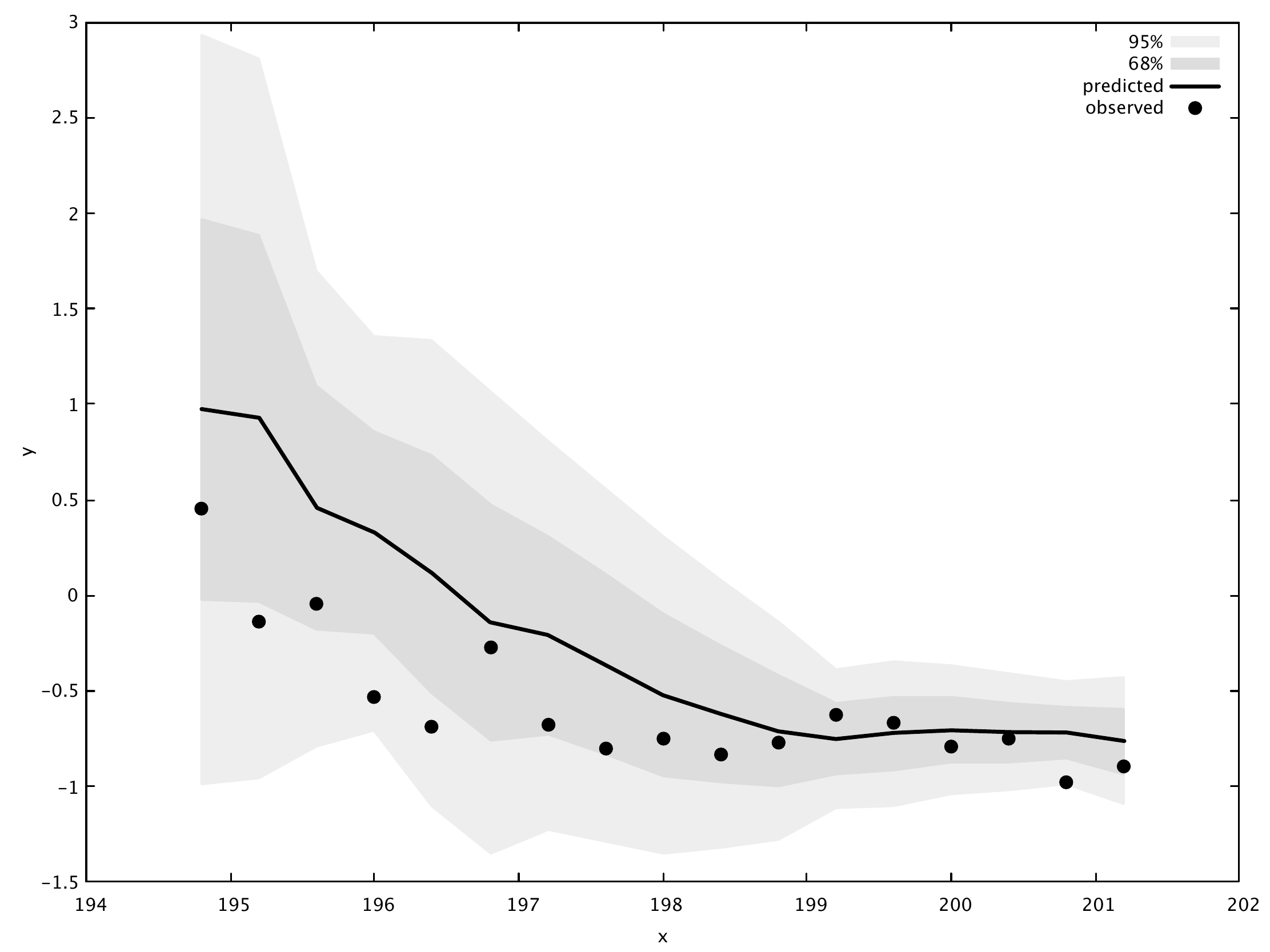}

		(c) deep GP, NLPD = -0.2183
    \end{minipage}
	\caption{Forecasting on Marathon dataset.}
	\label{fig:marathon-men-gold}
\end{figure*}

\begin{figure*}
  \begin{minipage}[c]{0.33\textwidth}
    \centering
	  \includegraphics[width=0.96\linewidth]{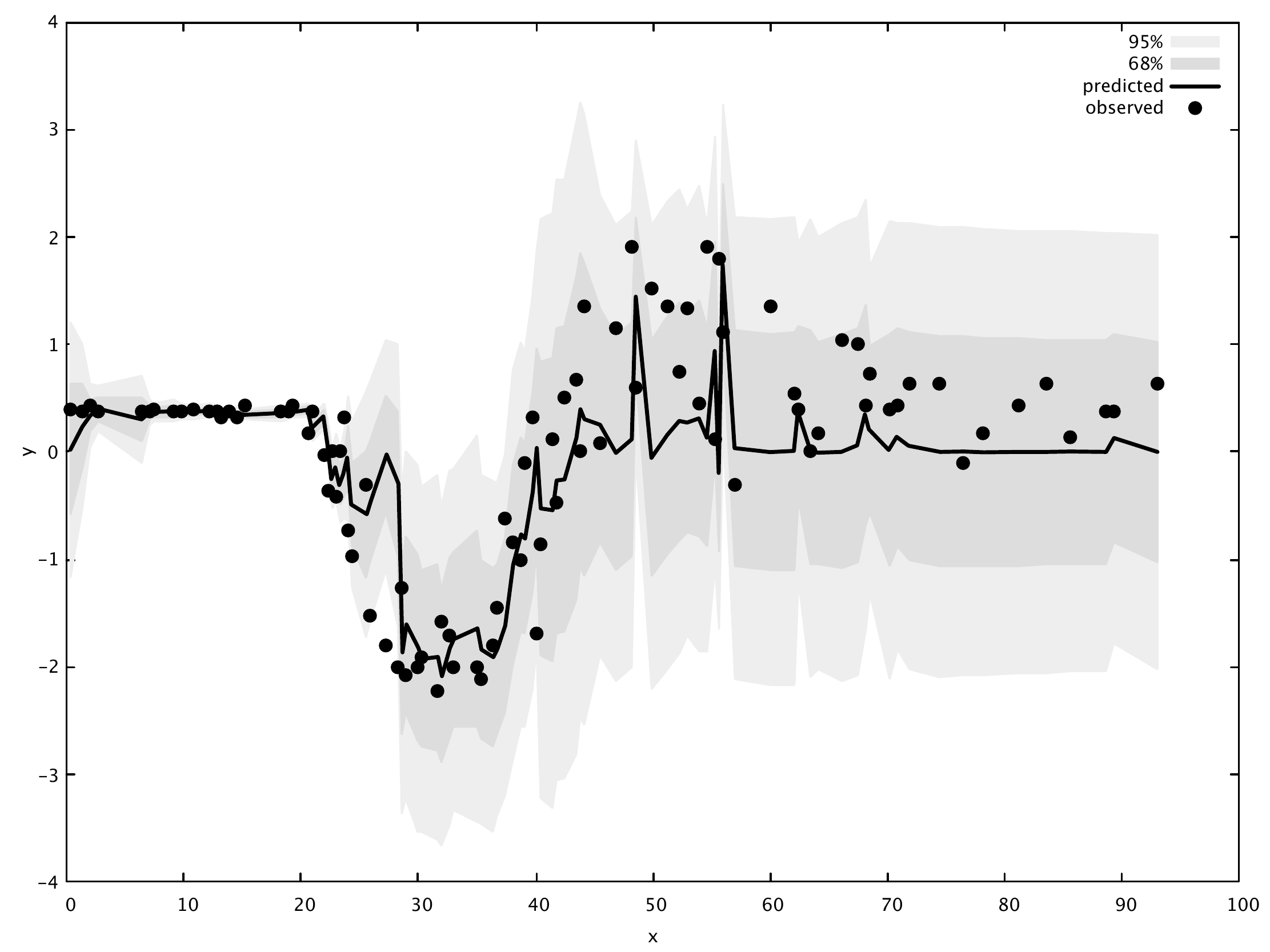}

		(a) no-warping, NLPD = 1.9320
    \end{minipage}
    \begin{minipage}[c]{0.33\textwidth}
    \centering
	\includegraphics[scale=0.5,width=0.96\linewidth]{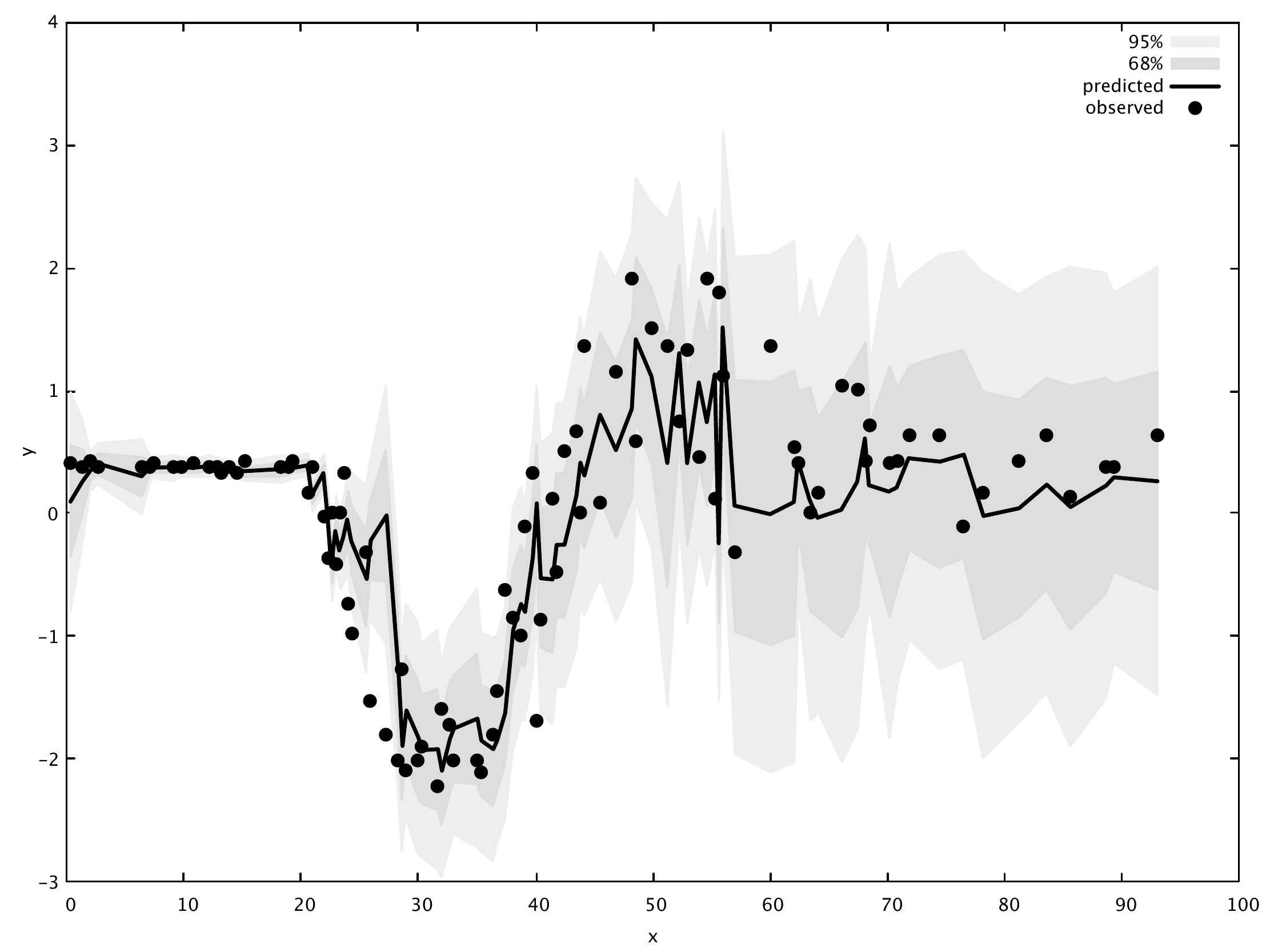}

		(b) warped, NLPD = 0.8063
    \end{minipage}
    \begin{minipage}[c]{0.33\textwidth}
    \centering
	\includegraphics[scale=0.5,width=0.96\linewidth]{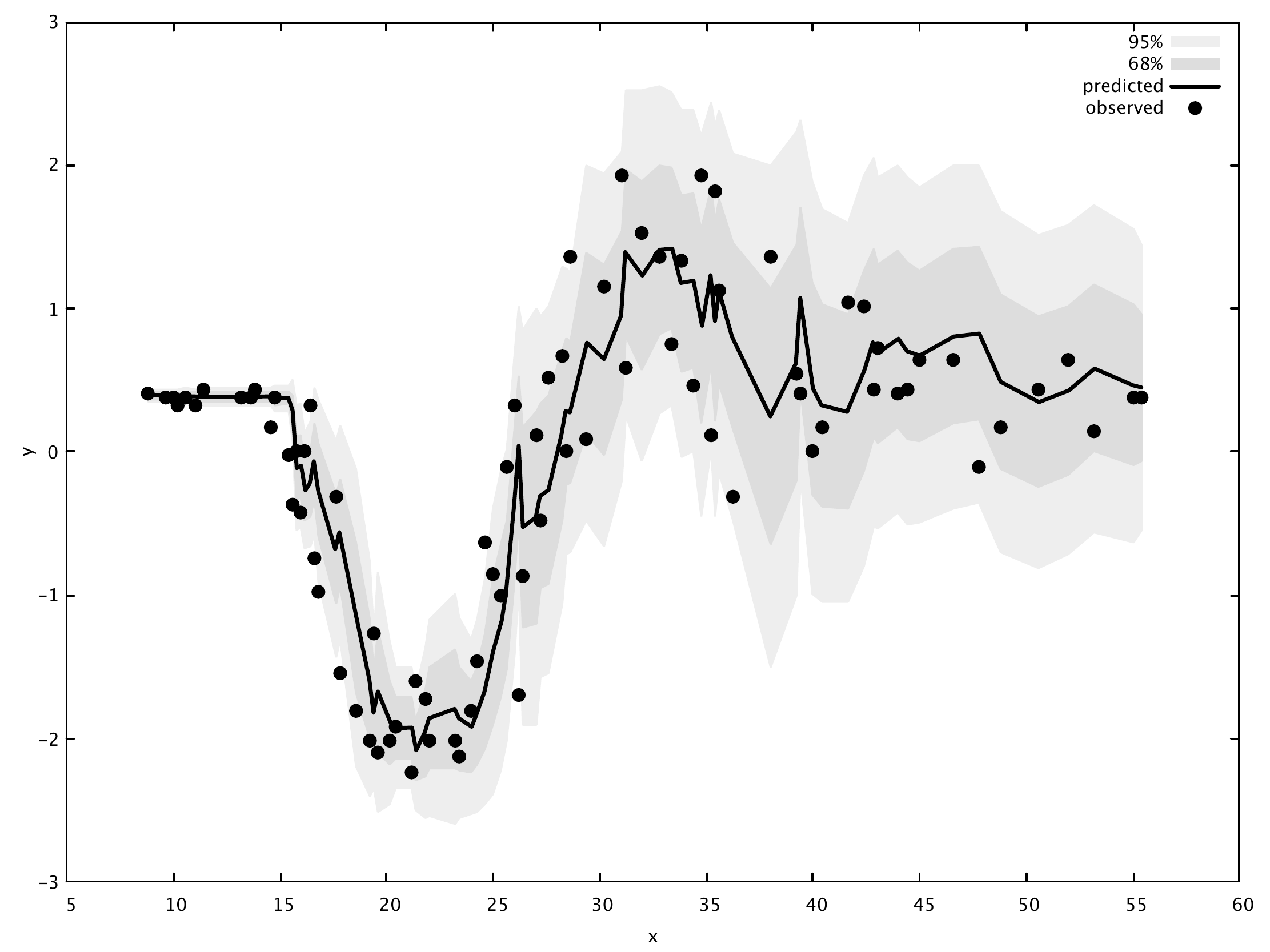}

		(c) deep GP, NLPD = 1.1919
    \end{minipage}
	\caption{Forecasting on Motorcycle dataset.}
	\label{fig:motor}
\end{figure*}

The empirical evaluation relies on modelling and inference
capabilities provided by differentiable probabilistic
programming~\cite{GW08,GHN+14}. We implemented WGP using
Infergo~\cite{T19}, a probabilistic programming facility, and
GoGP~\cite{GoGP}, a library for probabilistic programming with
Gaussian processes.  The source code, data, and detailed results
of empirical evaluations are available at
\url{https://bitbucket.org/dtolpin/wigp}. An implementation of
LBFGS~\cite{LN89} provided by Gonum~\cite{Gonum} was used 
for inferring hyperparameters. As a state-of-the-art algorithm
for non-stationary Gaussian process regression, we used an
implementation of deep Gaussian processes from
\url{https://github.com/SheffieldML/PyDeepGP}.

We evaluated the model on synthetic and real world data. Two
kernels were employed in the empirical evaluation:
\begin{enumerate}
	\item A $\mathrm{Matern}_{\frac 5 2}$ kernel, used with both
		synthetic and real-world data.
	\item A weighted sum of $\mathrm{Matern}_{\frac 5 2}$ kernel and a
		periodic kernel, used with synthetic data.
\end{enumerate}
The latter kernel was applied to synthetic data generated both
with and without seasonal component, to evaluate influence of
prior structural knowledge on one hand, and possible adverse
effect of model misspecification (periodic component where there
is no seasonality in the data) on the other hand. A
parameterized homoscedastic noise term was added to the kernel
in all evaluations. Vague log-normal priors were imposed on
kernel hyperparameters. 

\subsection{Synthetic datasets}

Synthetic data was generated by sampling 100 instances from
Gaussian processes with Matern(5/2) kernel, and with a sum of
Matern(5/2) and a periodic kernel, to emulate seasonality in the
data. To emulate non-stationarity, log distances between inputs
were sampled from a Gaussian process with an RBF kernel and then
unwarped into equidistant inputs. Samples from the periodic
kernel component were drawn for equidistant inputs, in
accordance with the assumption that the seasonality period is
fixed.

Table~\ref{tab:nlpd-synthetic} provides negative log
predictive density (NLPD) for regular, unwarped, Gaussian
process, warped Gaussian process with and without the
periodic component, and deep Gaussian process on
the synthetic dataset.  Smaller NLPD means better forecasting
accuracy. WGP outperforms both regular and deep Gaussian process
by a wide margin on the synthetic dataset. Using a kernel with
periodic component on seasonal data improves forecasting, but
accounting for non-stationarity through warping always results
in much better accuracy.

Figure~\ref{fig:synthetic} shows a typical forecast by each of
the models on a single instance from the synthetic dataset.

\subsection{Real-world datasets}

We used three real-world datasets for the evaluation:
\begin{itemize}
	\item Marathon --- olympic marathon time records for years
		1896--2016, obtained from
		\url{https://www.kaggle.com/jayrav13/olympic-track-field-results}.
	\item LIDAR --- observations from light detection and
		ranging experiment~\cite{S94}.
	\item Motorcycle --- data from a simulated motorcycle
		accident~\cite{S85}.
\end{itemize}

Table~\ref{tab:nlpd-real} compares performance of regular
Gaussian process WGP, and deep Gaussian process on the data
sets. WGP shows the best predictive performance on LIDAR and
Motorcycle data. On the Marathon time series, deep Gaussian
process performs slightly better, apparently due to smoothness
of the data. Figures~\ref{fig:marathon-men-gold}
and~\ref{fig:motor} show forecasting with each of the models on
Marathon and Motorcycle datasets. 

\section{RELATED WORK}

Work related to this research is concerned with Gaussian
processes for time series forecasting, non-stationarity in
Gaussian process regression, and warping of the input space for
representing non-stationarity, in the order of narrowing focus.
\cite{ROE+13} gives an introduction to Gaussian processes for
time series modelling, including handling of non-stationarity
through change point detection. 

Non-stationarity in Gaussian processes is attributed to either
heteroscedasticity, that is, varying observation noise, or to
non-stationarity of the covariance, or to both.
Heteroscedasticity is addressed by modelling dependency of noise
on the input and, possibly,
output~\cite{GWB98,LSC05,KPB07,WN12,DTL16}. Non-stationarity of
the covariance is represented through change
points~\cite{GOR09,STR10,CV11}, non-stationary
kernels~\cite{G97,PS03,PKB08} or input and output space
transformations (warping)~\cite{SG92,CPR+16,MG18,DL13,D18}.

Current work uses warping of the input space to represent
non-stationarity. However, unlike previous research, only
observation inputs are transformed rather than the whole
input space, allowing for a simpler representation and 
more efficient inference. Due to the non-parametric nature of
transformation employed in this work, the introduced model
is applicable to time series both with change points and  with
smooth non-stationarities.

\section{CONCLUSION}

We introduced a Gaussian process-based model where
non-stationarity is handled through non-parametric warping of
observation inputs. In application to time series, the model
facilitates forecasting of future observations with variances
depending on outputs, as well as inputs, of past observations,
while staying within the framework of `standard' Gaussian
process inference. When the data is known to possess periodic
properties or non-local correlations, these correlations can be
encoded in the model while still handling non-stationarity
through warping. The introduced approach to input warping can be
used with existing Gaussian process libraries and algorithms,
and there is room for compromise between accuracy of modelling
non-stationarity and computation time.

It still remains an open question to which extent a more expressive
warping may improve the quality of forecasting. Combining the
introduced model with change-point detection may be beneficial
in cases of abrupt changes in process parameters. Still, in
cases where simplicity of implementation and robustness in face
of variability in time series are of high priority, the
introduced model appears to provide a practical and efficient
solution.

\clearpage
\bibliography{refs}

\end{document}